\def\year{2019}\relax
\documentclass[letterpaper]{article}

\usepackage{aaai19}
\usepackage{times}
\usepackage{helvet}
\usepackage{courier}
\usepackage[hyphens]{url}
\usepackage{graphicx}

\usepackage{amsfonts}
\usepackage{bm}
\usepackage{amsmath}
\usepackage{booktabs}
\usepackage{bbm}
\usepackage{mathtools}
\usepackage{tabularx}
\usepackage{textcomp}
\usepackage{subcaption}

\newcommand{\citet}[1]{\citeauthor{#1} \shortcite{#1}} 
\newcommand{\citep}{\cite} 
\newcommand{\citealp}[1]{\citeauthor{#1} \citeyear{#1}}

\urldef{\mycodeurl}\url{https://github.com/DevSinghSachan/ssl_text_classification}
\urldef{\corenlpurl}\url{https://stanfordnlp.github.io/CoreNLP/}
\urldef{\parelexturl}\url{https://github.com/yuhaozhang/tacred-relation}

\DeclarePairedDelimiterX{\infdivx}[2]{(}{)}{%
  #1\;\delimsize\|\;#2%
}
\newcommand{\infdiv}{D_{\textit{KL}}\infdivx}

\title{Revisiting LSTM Networks for Semi-Supervised \\Text Classification via Mixed Objective Function}
\author{Devendra Singh Sachan\\
Petuum Inc\\
{sachan.devendra@gmail.com}\\
\And
Manzil Zaheer\\
Google Research\\
{manzilzaheer@google.com}
\And 
Ruslan Salakhutdinov\\
Carnegie Mellon University\\
{rsalakhu@cs.cmu.edu}
}

\begin{document}
\maketitle

\begin{abstract}
\begin{quote}
    In this paper, we study bidirectional LSTM network for the task of text classification using both supervised and semi-supervised approaches. Several prior works have suggested that either complex pretraining schemes using unsupervised methods such as language modeling~\citep{dai2015semi,miyato2016adversarial} or complicated models~\citep{johnson2017deep} are necessary to achieve a high classification accuracy. However, we develop a training strategy that allows even a simple BiLSTM model, when trained with cross-entropy loss, to achieve competitive results compared with more complex approaches. Furthermore, in addition to cross-entropy loss, by using a combination of entropy minimization, adversarial, and virtual adversarial losses for both labeled and unlabeled data, we report state-of-the-art results for text classification task on several benchmark datasets. In particular, on the ACL-IMDB sentiment analysis and AG-News topic classification datasets, our method outperforms current approaches by a substantial margin. We also show the generality of the mixed objective function by improving the performance on relation extraction task.\footnote{\mycodeurl}
\end{quote}
\end{abstract}

\section{Introduction}
Text classification is an important problem in natural language processing (NLP). The task is to assign a document to one or more predefined categories. It has a wide range of applications such as sentiment analysis~\citep{pang2008opinion}, topic categorization~\citep{lewis2004rcv1}, and email filtering~\citep{sahami98bayesian}. Early machine learning approaches for text classification were based on the extraction of bag-of-words features followed by a supervised classifier such as na\"ive Bayes~\citep{mccallum1998comparison} or a linear SVM~\citep{joachims1998text}. Later, better word representations were introduced, such as latent semantic analysis~\citep{Deerwester90indexingby}, skipgram~\citep{mikolov2013distributed}, and fastText~\citep{joulin2017fastText}, which improved classification accuracy. Recently, recurrent and convolutional neural network~\citep{kim2014convolutional} models were introduced to utilize the word order and grammatical structure. Many complex variations of these models have been proposed to improve the text classification accuracy, e.g. training one-hot CNN (JZ15a;~\citealp{johnson2015effective}) or one-hot bidirectional LSTM (BiLSTM) network with dynamic max-pooling (JZ16;~\citealp{johnson2016supervised}).

Current state-of-the-art approaches for text classification involve using pretrained LSTMs (DL15; \citealp{dai2015semi}) or complex computationally intensive models (JZ17; \citealp{johnson2017deep}). DL15 argued that randomly initialized LSTMs are difficult to optimize and can lead to worse performance than linear models. Therefore, to improve the performance, they proposed \emph{pretraining} the LSTM with either a language model or a sequence auto-encoder. However, pretraining or using complicated models can be very time consuming, which is a major disadvantage and may not be always feasible. In this paper, we consider a BiLSTM classifier model similar to the one proposed by DL15 for text classification. For this simple BiLSTM model with pretrained embeddings, we propose a training strategy that can achieve accuracy competitive with the previous purely supervised models, but without the extra pretraining step. We also perform ablation studies to understand aspects of the proposed training strategy that result in an improvement.

Pretraining approaches often use extra unlabeled data in addition to the labeled data. 
We explore the applicability of such semi-supervised learning (SSL) in our training framework, where there is no prior pretraining step.
In this regard, we propose a \emph{mixed objective function} for SSL that can utilize both labeled and unlabeled data to obtain further improvement in classification. To summarize, our contributions are as follows:
\begin{itemize}
\itemsep0em
\item We show that with proper model training, using a maximum likelihood objective with a simple one-layer BiLSTM model (\S\ref{sec:methods}) can produce competitive accuracies,
\item We propose a mixed objective function that can be applied to text classification tasks (\S\ref{sec:methods}),
\item On seven benchmark text classification tasks, we achieve new state-of-the-art results despite having a much simpler model, minimal model tuning, and fewer parameters (\S\ref{sec:results}),
\item We extend our proposed mixed objective function to relation extraction task, where we achieve better F1 score on SemEval-2010 and TACRED datasets, again with a simple model and minimal model tuning (\S\ref{sec:RE}).
\end{itemize}

\section{Methods} \label{sec:methods}
In this section, we will describe the model architecture, our training strategy, and our proposed mixed objective function. 
For mathematical notation, we will use bold lowercase to denote vectors, bold uppercase to denote matrices, and lowercase to denote scalars and individual words in a document.

\subsection{Model Architecture}
\begin{figure}[t]
\centering
\includegraphics[scale=0.85]{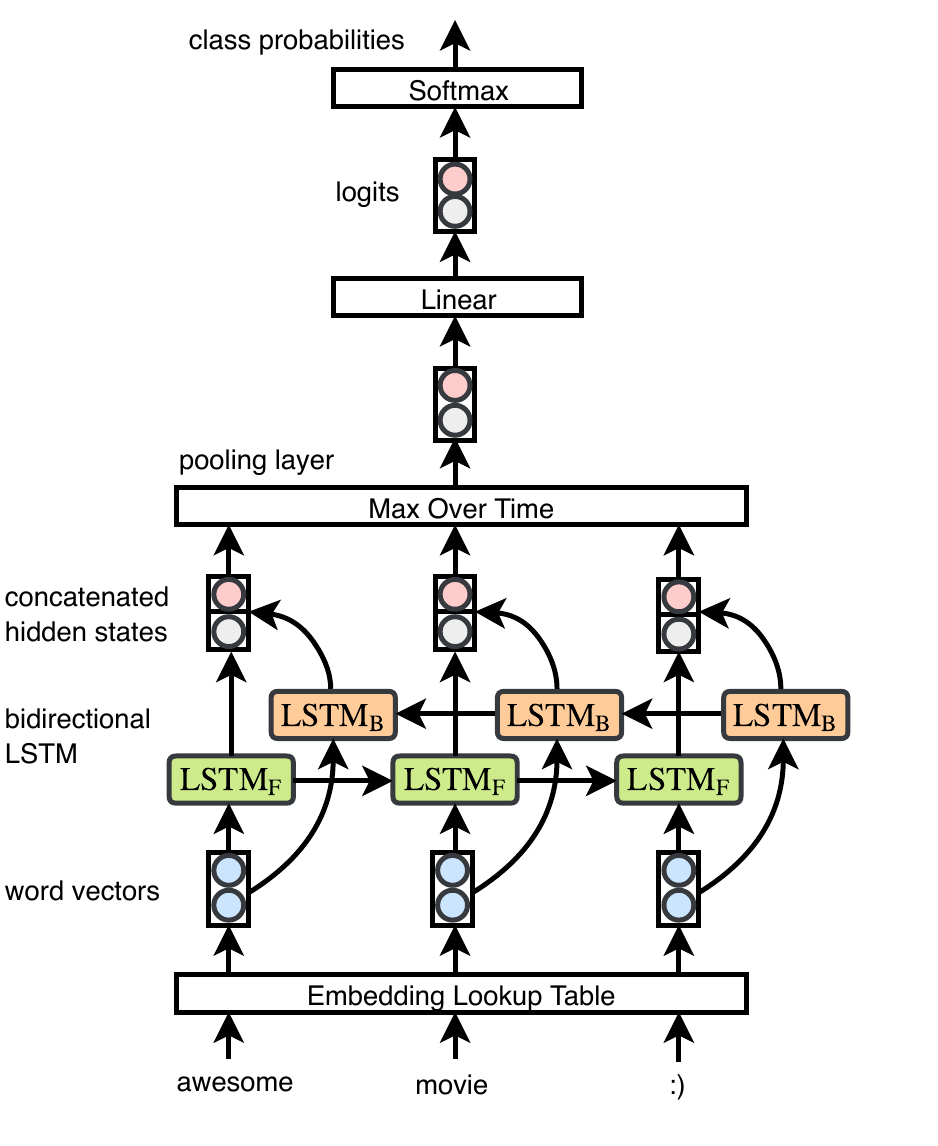}
\caption{Text classification model architecture.}
\label{fig:model}
\end{figure}

The classification model consists of an embedding layer, a bidirectional long short-term memory (BiLSTM) encoder~\citep{hochreiter1997long,schuster1997bidirectional}, a max-pooling layer, a linear (fully-connected) layer, and finally a softmax layer (see Figure~\ref{fig:model}). 
First, the sequence of words ($w_1, \ldots, w_{\textit{T}}$) contained in a document are passed through an embedding layer that contains a lookup table which maps them to dense vectors 
($\mathbf{v}_1,\ldots,\mathbf{v}_{\textit{T}},\;\mathbf{v}_t\in \mathbb{R}^d$). 
Next, the forward and backward LSTM of the BiLSTM encoder processes these word vectors in the forward (left to right) and backward (right to left) directions respectively, updating corresponding hidden states at each time-step
\begin{align*}
\overrightarrow{\mathbf{h}_t} = \overrightarrow{\mathrm{LSTM_F}}(\overrightarrow{\mathbf{h}_{t-1}}, \mathbf{v}_t),\;\overleftarrow{\mathbf{h}_t} = \overleftarrow{\mathrm{LSTM_B}}(\overleftarrow{\mathbf{h}_{t-1}}, \mathbf{v}_t).
\end{align*}
Next, these hidden state outputs from the forward LSTM ($\overrightarrow{\mathbf{h}_t}$) and backward LSTM ($\overleftarrow{\mathbf{h}_t}$) are concatenated at every time-step to enable encoding of information from past and future contexts respectively
\begin{align*}
\mathbf{h}_t &= [\overrightarrow{\mathbf{h}_t} \;||\;\overleftarrow{\mathbf{h}_t}],\;t \in [1, T],\;\mathbf{h}_t\in\mathbb{R}^n.
\end{align*}
These concatenated hidden states are next fed to the pooling layer that computes the maximum value over time to obtain the feature representation of the input sequence ($\mathbf{h} \in \mathbb{R}^n$) 
\begin{align*}
h^{\ell} = \max_{t} h^{\ell}_{t},\;t \in [1, T],\;\forall{\ell}\in\{1,\ldots,n\}.
\end{align*}
This max-pooling mechanism constrains the model to
capture the most useful features produced by the BiLSTM encoder. Next, the linear layer applies an affine transformation to the feature vector to produce \emph{logits} ($\mathbf{d}$)
\begin{align*}
\mathbf{d} = \mathbf{W}\mathbf{h} + b,\;\mathbf{d} \in \mathbb{R}^\textit{K},
\end{align*}
where $\textit{K}$ is the number of classes, $\mathbf{W}$ is the weight matrix, and $b$ is the bias. Next, these logits are normalized using the $\mathrm{softmax}$ function to give our estimated class probabilities as
\begin{align*}
p(y=k|\mathbf{x}; \boldsymbol{\theta}) = \frac{\exp(d_{k})}{\sum_{j=1}^{\textit{K}} \exp(d_{j})},\;\forall{k}\in\{1,\ldots,\textit{K}\},
\end{align*}
where $(\mathbf{x}, y)$ is a training example and $\boldsymbol{\theta}$ denotes the model parameters.
For model training, we use supervised and unsupervised loss functions, which are discussed next.

\subsection{Supervised Training}
Let there be $\textit{m}_\ell$ labeled examples in the training set that are denoted as $\{(\mathbf{x}^{(1)},y^{(1)}),\ldots,(\mathbf{x}^{(\textit{m}_\ell)},y^{(\textit{m}_\ell)})\}$, where $\mathbf{x}^{(i)}$ represents a document's word sequence, $y^{(i)}$ represents class label such that $y^{(i)}\in\{1,2,\ldots,K\}$. For supervised training of the classification model, we make use of two methodologies: \emph{maximum likelihood} estimation and \emph{adversarial training}, which are described next.

\subsubsection*{\textbf{Maximum Likelihood (ML)}} This is the most widely used method to learn the parameters of neural network models from observed data for text classification task. Here, we minimize the average cross-entropy loss between the estimated class probability and the ground truth class label for all training examples
\begin{align*}
L_{\textit{ML}}(\boldsymbol\theta) = \frac{-1}{m_{\ell}}\sum_{i=1}^{m_{\ell}} \sum_{k=1}^{\textit{K}} \mathbbm{1}(y^{(i)}=k)\log{p(y^{(i)}=k|\mathbf{x}^{(i)}; \boldsymbol{\theta})},
\end{align*}
where $\mathbbm{1}(;)$ is an indicator function.

\subsubsection*{\textbf{Adversarial Training (AT)}} Adversarial examples are created from inputs by small perturbations to mislead the machine learning algorithm. The objective of adversarial training is to construct and give as input adversarial examples during model training procedure to make the model more robust to adversarial noise and thereby improving its generalization ability~\citep{goodfellow2014explaining}. 

In this work, we make adversarial perturbations to the input word embeddings ($\mathbf{v} = [\mathbf{v}_1,\ldots,\mathbf{v}_T]$) (MDG16,~\citealp{miyato2016adversarial}). These perturbations ($\mathbf{r}_{\textit{at}}$) are estimated by linearizing the supervised cross-entropy loss around the input word embeddings. Specifically, to get the adversarial embedding $(\mathbf{v}^*)$ corresponding to $\mathbf{v}$, we use the $L_2$ norm of the training loss gradient ($\mathbf{g}$) that is computed by backpropagation using the current model parameters ($\boldsymbol{\hat{\theta}}$)
\begin{align*}
\mathbf{r}_{\textit{at}} &= \epsilon \mathbf{g}/\|\mathbf{g}\|_2,\;\text{where}\;\mathbf{g} = -\nabla_{\mathbf{v}}\log p(y=k|\mathbf{v};\boldsymbol{\hat{\theta}})\\
\mathbf{v}^{*} &= \mathbf{v} + \mathbf{r}_{\textit{adv}}
\end{align*}
where, $k$ is the correct class label, $\epsilon$ is a hyperparameter that controls the magnitude of the perturbation. We apply adversarial loss to only the labeled data. It is defined as
\begin{align*}
L_{\textit{AT}}(\boldsymbol{\theta}) = \frac{-1}{m_{\ell}}
\sum_{i=1}^{m_{\ell}}\sum_{k=1}^{K} \mathbbm{1}(y^{(i)}=k)\log p(y^{(i)}=k|\mathbf{v}^{*(i)}; \boldsymbol{\hat{\theta}}).
\end{align*}

\subsection{Unsupervised Training}
In this paper, in addition to supervised training, we also experiment with two unsupervised methodologies: \emph{entropy minimization} and \emph{virtual adversarial training}.  These loss functions when incorporated into the objective function act as effective regularizers during model training. To describe them, we assume that there exists an additional $\textit{m}_\textit{u}$ unlabeled examples in the dataset $\{\mathbf{x}^{(1)},\ldots, \mathbf{x}^{(\textit{m}_{\textit{u}})}\}$.

\subsubsection{\textbf{Entropy Minimization (EM)}} In addition to supervised cross-entropy loss, we also minimize the conditional entropy of the estimated class probabilities~\citep{grandvalet2005semi,miyato2017virtual}. This can also be interpreted as a special case of the missing label problem where the probability $p(y^{(i)}=k|\mathbf{x}^{(i)};\boldsymbol{\theta})$ signifies a soft assignment of the $i^{th}$ example to label $k$ (\emph{i.e.}\ soft clustering). Entropy minimization loss is applied in an unsupervised manner to both the labeled and unlabeled data.
\begin{equation*}
L_{\textit{EM}}(\boldsymbol{\theta}) = \frac{-1}{m} \sum_{i=1}^{m} \sum_{k=1}^{\textit{K}} p(y^{(i)}=k|\mathbf{x}^{(i)})\log p(y^{(i)}=k|\mathbf{x}^{(i)}),
\end{equation*}
where $m = {m_{\ell} + m_{\textit{u}}}$ and dependence on $\boldsymbol{\theta}$ is suppressed.

\begin{table}[b]
\small
\centering
\begin{tabular}{@{}l|r r r r@{}}
 \toprule
 \textbf{Dataset} & \multicolumn{1}{c}{\textbf{Train}} & \multicolumn{1}{c}{\textbf{Test}} & \textit{\textbf{K}} & \multicolumn{1}{c}{$\boldsymbol{\ell}$}\\
 \midrule
 ACL-IMDB & 25,000 & 25,000 & 2 & 268 \\
 Elec & 25,000 & 25,000 & 2 & 125 \\
 AG-News & 120,000 & 7,600 & 4 & 46 \\
 DBpedia & 560,000 & 70,000 & 14 & 56 \\
 RCV1 & 15,564 & 49,821 & 51 & 120 \\
 IMDB & 1,490,000 & 1,070,000 & 5 & 280 \\
 Arxiv & 664,000 & 443,000 & 127 & 153 \\
 \bottomrule
 \end{tabular}
\caption{Summary statistics for text classification datasets; \textit{K} = number of classes; \textit{$\ell$} = average length of a document.}
\label{table:dataset_stat}
\end{table}

\subsubsection{\textbf{Virtual Adversarial Training (VAT)}} 
As opposed to minimizing the cross-entropy loss of the adversarial examples in AT, in VAT, we minimize the KL divergence between $p(\mathbf{v})$ and $p(\mathbf{v}^*)$, where $\mathbf{v}^* = \mathbf{v} + \mathbf{r}_{\textit{vat}}$. The motivation of using KL divergence as an additional loss term in the objective function is that it tends to make the loss surface smooth at the current example~\citep{miyato2017virtual}. Also, computing the VAT loss doesn't require class labels, so it can be applied to unlabeled data as well. In this work, we follow the approach proposed by MDG16 that makes use of the second-order Taylor expansion of distance followed by power iteration method to approximate the virtual adversarial perturbation. First, an i.i.d. random unit vector is sampled for every example from the Normal distribution ($\mathbf{d}^{(i)}\sim\mathcal{N}(\mathbf{0},\:\mathbf{I})\in\mathbb{R}^d$) and then adversarial perturbation computed as $\xi \mathbf{d}^{(i)}$ is added to the word embeddings, where $\xi$ is a hyperparameter
\begin{align*}
\mathbf{v}'^{(i)} = \mathbf{v}^{(i)} + \xi\mathbf{d}^{(i)}.
\end{align*}
Next, the gradient is estimated from the KL divergence as:
\begin{align*}
\mathbf{g} = \nabla_{\mathbf{v'}}\infdiv
{p(. \mid \mathbf{v}^{(i)}; \boldsymbol{\hat{\theta}})}
{p(. \mid \mathbf{v}'^{(i)}; \boldsymbol{\hat{\theta}})}.
\end{align*}
Virtual adversarial perturbation ($\mathbf{r}_{\textit{vadv}}$) is generated using the $L_2$ norm of the gradient and added to the word embeddings
\begin{align*}
\mathbf{r}^{(i)}_{ \textit{vat} } &= \epsilon \mathbf{g}/\|\mathbf{g}\|_2\\
\mathbf{v}^{*(i)} &= \mathbf{v}^{(i)} + \mathbf{r}^{(i)}_{\textit{vat}}\;.
\end{align*}
Lastly, virtual adversarial loss can be computed from both the labeled and unlabeled data as:
\begin{align*}
L_{\textit{VAT}}(\boldsymbol{\theta}) = \frac{1}{m} \sum_{i=1}^{m} \infdiv{p(. \mid \mathbf{v}^{(i)};\boldsymbol{\theta})}{p(. \mid \mathbf{v}^{*(i)};\boldsymbol{\theta})},
\end{align*}
where $m = m_{\ell} + m_\textit{u}$.

\subsection{Mixed Objective Function}
Our proposed mixed objective function combines the above described supervised and unsupervised loss functions using $\lambda_{\textit{ML}}$, $\lambda_{\textit{AT}}$, $\lambda_{\textit{EM}}$, and $\lambda_{\textit{VAT}}$ as hyperparameters
\begin{align*}
L_{\textit{MIXED}} = \lambda_{\textit{ML}} L_{\textit{ML}} + \lambda_{\textit{AT}}L_{\textit{AT}} + \lambda_{\textit{EM}}L_{\textit{EM}} + \lambda_{\textit{VAT}} L_{\textit{VAT}}.
\end{align*}

\begin{table}[b]
\small
\centering
\begin{tabular}{@{}l | r r r@{}}
 \toprule
 \textbf{Dataset} & \multicolumn{1}{c}{\textbf{BSize}} & \multicolumn{1}{c}{\textbf{Vocab}} & $\boldsymbol{\epsilon}$ \\
 \midrule
 ACL-IMDB & 3,000 & 80,000 & 5 \\
 Elec & 2,000 & 40,000 & 2 \\
 AG-News & 2,000 & 75,000 & 1 \\
 DBpedia & 7,500 & 50,000 & 1 \\
 RCV1 & 2,000 & 100,000 & 2 \\
 IMDB & 15,000 & 150,000 & 5 \\
 Arxiv & 8,000 & 100,000 & 1 \\
 \bottomrule
 \end{tabular}
\caption{Dataset-specific hyperparameters; BSize = number of tokens in a minibatch; Vocab = number of words present in vocabulary; $\epsilon$ = adversarial perturbation.}
\label{table:hyperparameters}
\end{table}

\subsection{Training Strategy}
We note that there are three main considerations to be taken into account when training the text classifier.
First, the knowledge of the entire sequence is essential for good classification performance. Thus, commonly used practice of truncated backpropagation through time~\citep{werbos1988generalization} is a key limiting factor. One should perform gradient update for the entire text sequence. To prevent out of memory issues that can result from longer sequences, we propose to use \emph{dynamic batch size} that consists of fixed number of total words per mini-batch. Second, not only we need to use pretrained word embeddings, but we need to finetune them for the specific task. Lastly, we should use a larger vocabulary size and not limit to only high-frequency words. This is because rare or tail words are often strong indicators of the class.

\begin{table*}[!t]
\centering
\small
\begin{tabular}{@{}l | r r r r r r r@{}}
\toprule
\textbf{Model} & \textbf{ACL-IMDB} & \textbf{Elec} & \textbf{AG-News} & \textbf{DBpedia} & \textbf{RCV1} & \textbf{IMDB} & \textbf{Arxiv} \\
\midrule
Linear Model (TFIDF + SVM) & 9.51 & 9.16 & 7.64 & 1.31 & 10.68 & 40.00 & 34.81 \\
Vanilla LSTM [\citet{dai2015semi}] & 10.00 & -- & -- & 13.64 & -- & -- & -- \\
DocVec [\citet{le2014distributed}] & 11.40 & 8.95 & 8.00 & 2.34 & -- & 44.10 & 37.40 \\
FastText [\citet{joulin2017fastText}] & 11.34 & -- & 7.50 & 1.40 & -- & 42.13 & 33.23 \\
CNN [\citet{kim2014convolutional}] & 9.17 & 8.03 &  5.92 & 0.98 & 10.44 & 49.53 & 34.21 \\
oh-CNN [Johnson and Zhang~\shortcite{johnson2015semi,johnson2017deep}] & 7.67 & \textbf{7.14} & 6.88 & \textbf{0.88} & 9.17 & 38.15 & 35.89 \\
char-CNN [\citet{zhang2015character}] & -- & -- & 9.51 & 1.55 & -- & -- & -- \\
\midrule
$L_{\textit{ML}}$ [Our Method] & \textbf{6.43} & 7.40 & \textbf{5.62} & 0.91 & \textbf{7.78} & \textbf{35.64} & \textbf{31.76} \\
\bottomrule
\end{tabular}
\caption{Error rates (\%) when the model is trained using $L_{\textit{ML}}$ and comparison with previous best supervised methods.}
\label{table:accuracy}
\end{table*}

\begin{table*}[t]
\small
\centering
\begin{tabular}{@{}l | r r r r r r r@{}}
\toprule
\textbf{Model} & \textbf{ACL-IMDB} & \textbf{Elec} & \textbf{AG-News} & \textbf{DBpedia} & \textbf{RCV1} & \textbf{IMDB} & \textbf{Arxiv} \\
\midrule
SA-LSTM [\citet{dai2015semi}] & 7.24 & -- & -- & 1.19 & 7.40 & -- & -- \\
LSTM [\citet{miyato2016adversarial}] & 5.91 & 5.40 & 6.78 & 0.76 & 6.68 & 35.85 & 30.97 \\
oh-LSTM [Johnson and Zhang~\shortcite{johnson2016supervised,johnson2017deep}] & 5.94 & 5.55 & 6.57 & 0.84 & 7.15 & 37.56 & 31.17 \\
ULMFit [\citet{howard2018universal}] & 4.60 & -- & 5.01 & 0.80 & -- & -- & -- \\
\midrule
$L_{\textit{MIXED}}$ [Our Method] & \textbf{4.32} & \textbf{5.24} & \textbf{4.95} & \textbf{0.70} &  \textbf{6.23} & \textbf{34.04} & \textbf{29.95} \\
\bottomrule
\end{tabular}
\caption{Error rates (\%) when the model is trained using $L_{\textit{MIXED}}$ and comparison with previous best semi-supervised methods.}
\label{table:accuracy_ssl}
\end{table*}

\section{Experiments}\label{sec:exp_setup}
\subsection*{Dataset Description}
In this work, we experiment with seven datasets that are summarized in Table~\ref{table:dataset_stat}. \emph{ACL-IMDB}~\citep{maas-EtAl:2011:ACL-HLT2011} and \emph{Elec} (JZ15a) datasets are widely used for binary sentiment classification of movie reviews and Amazon product reviews respectively while \emph{AG-News}, \emph{DBpedia}~\citep{zhang2015character}, and \emph{RCV1}~\citep{lewis2004rcv1} are for topic classification of news articles, Wikipedia, and Reuters corpus respectively. For the RCV1 dataset, we perform multiclass topic classification based on the second-level topics and construct its training, dev, and test splits in accordance with JZ15a. To show that our proposed method also scales to larger datasets and categories, we also experiment with large-scale datasets of \emph{IMDB} reviews and \emph{Arxiv} abstracts that are used for fine-grained sentiment- and topic classification respectively~\citep{sachan2018investigating}. We preprocess all the datasets by converting the text to lowercase and treat all punctuations as separate tokens.

\subsection*{Implementation Details}
All of our models were implemented in PyTorch framework~\citep{paszke2017automatic} and were trained on a single GPU. Our experimental setup is common for all the datasets unless specified otherwise. We use 300D pretrained vectors to initialize the embedding layer. We learn embeddings for the ACL-IMDB, RCV1, IMDB, and Arxiv datasets using \verb+word2vec+~\citep{mikolov2013distributed} and use \verb+fastText+ pretrained embeddings\footnote{These embeddings were trained on data containing 600B tokens from Common Crawl (\emph{crawl-300d-2M.vec}).}~\citep{mikolov2018advances} for all the other datasets. We use one-layer BiLSTM of size 512D. For regularization, we apply dropout ($p_{drop}=0.5$) to word embeddings and to LSTM's hidden states. We also use the word dropout strategy in which we randomly set a word to be ``UNK'' with a probability $p_w=0.1$. For training, we employ SGD using \emph{Adam} optimizer~\citep{DBLP:journals/corr/KingmaB14} ($\textit{learning rate}=10^{-3}$, $\beta_1=0$, $\beta_2=0.98$, $\epsilon_{\textit{adam}}=10^{-8}$) with an exponential learning rate decay scheme. We perform gradient clipping by having a maximum $L_2$ norm of 1. For training, we backpropagate through time over entire sequence, i.e. we did not truncate sequence. This differs  from DL15  where they perform truncated backpropagation through time for 400 time-steps from the end of a sequence.

For semi-supervised training, we experiment with all the objective functions described in~\S\ref{sec:methods}. For $L_{\textit{MIXED}}$, we include all the constituent terms with $\lambda_{\textit{ML}}$, $\lambda_{\textit{AT}}$, $\lambda_{\textit{EM}}$, $\lambda_{\textit{VAT}}$ set to 1 and $\xi=0.1$. We want to emphasize that in contrast with MDG16, we do not perform \emph{embedding layer normalization} during AT or VAT objectives, as by including it, we noticed a drop in accuracy during our initial experiments. We select the hyperparameters such as dynamic batch size, vocabulary size, and adversarial perturbation ($\epsilon$) by cross-validation on the development set. We mention these dataset-specific hyperparameters in Table~\ref{table:hyperparameters}. For supervised experiments ($L_{\textit{ML}}$), we perform training till 20 epochs and for semi-supervised experiments ($L_{\textit{MIXED}}$), training is done till 50 epochs. For ACL-IMDB, Elec, RCV1, IMDB, and Arxiv datasets, we use training and test set as unlabeled data, while for AG-News and DBpedia datasets as their test sets are small, we use only the training set as unlabeled data. During training, we keep the batch size of the unlabeled data the same as that of the labeled data.

\section{Results}
\label{sec:results}
In this section, we report the classification accuracy on the test set and perform ablation studies for both supervised and semi-supervised training.

\subsection{Maximum Likelihood Training}
In Table~\ref{table:accuracy}, we present the error rates of our method and the previous best-published models when training is done using only the maximum likelihood objective ($L_{\textit{ML}}$). We observe that our model that consists of one-layer BiLSTM and pretrained embedding weights achieves a very competitive performance on all the datasets compared with the more complex approaches such as one-hot LSTM (JZ16) or pyramidal CNN (JZ17). Specifically, for ACL-IMDB, AG-News, IMDB, and Arxiv datasets, we report much better results than earlier methods. Thus, our proposed model and training strategy enjoy the following advantages: (a) it is very easy to implement using current deep learning frameworks; (b) it requires much less training time and GPU memory compared with other complicated models; (c) it entirely avoids complex initialization strategies such as pretraining the LSTM weights using a language model; (d) Our results can serve as strong baselines when developing more advanced task-specific models.

\begin{table}[t]
\small
\centering
\resizebox{\linewidth}{!}{
\begin{tabular}{@{}c c c c c c | c@{}}
\toprule
$N$ & $\textit{Embedding}$ & $p_{\textit{w}}$ & $\textit{BSize}$ & $H$ & $\textit{Vocab}$ & ACL-IMDB \\
\midrule
1 & $\textit{Finetune}$ & 0.1 & 3,000 & 512 & 80,000 & 6.43 \\
\midrule
2 &          &     &       &     &        & 6.47 \\
% 3 &          &     &       &     &        & 7.13 \\
  & $\textit{Random}$   &     &       &     &        & 7.64 \\
  & $\textit{Static}$ &     &         &      &       & 8.17 \\
  &   & 0.0 &         &      &       & 6.57 \\
  &   &     & 1,000 &      &       & 6.98 \\
  &   &     &         & 256 &        & 6.67 \\
  &   &     &         & 1,024 &        & 7.05 \\
  &   &     &         &     & 30,000 & 7.78 \\
\midrule
\multicolumn{6}{c|}{$\textit{No text preprocessing}$} & 8.80 \\
\bottomrule
\end{tabular}}
\caption{Error rates (\%) of variations on the BiLSTM model trained using $L_{\textit{ML}}$ on the ACL-IMDB dataset. Unlisted values are identical to those of the first row; $N$ = number of BiLSTM layers; $H$ = LSTM hidden size.}
\label{table:ablation}
\end{table}

To know the importance of various components in the model and training regimen, we perform ablation studies using the ACL-IMDB dataset (see Table~\ref{table:ablation}). We verify that good performance of our model mostly results from finetuning the pretrained embeddings, using a larger vocabulary size, and using a carefully preprocessed dataset. We also see that excluding word dropout, smaller-sized LSTM, and lowering the batch size causes a slight drop in performance while using static pretrained or randomly initialized embeddings or smaller vocabulary size can cause a large drop.

\subsection{Semi-Supervised Training}
For our next set of experiments, we perform training using $L_{\textit{MIXED}}$ objective whose results are shown in Table~\ref{table:accuracy_ssl}. We observe that the mixed objective improves over maximum likelihood objective and achieves state-of-the-art results on all the seven datasets. Specifically, on the widely used ACL-IMDB dataset, there is a substantial reduction of 26.9\% in relative error compared with the previous best-published model of JZ16, which was substantially more complex as they use one-hot encodings of words along with a lot of additional features such as multi-view region embeddings from CNNs and LSTMs. We also want to highlight that, although the model of MDG16 also experiments with adversarial and virtual adversarial training, our approach performs much better compared with them due to our improved training strategy and the use of $L_{\textit{EM}}$ objective. Similarly, for the benchmark AG-News dataset, we observe relative error reduction of 26.6\% compared with previous state-of-the-art model of JZ17 who use a very deep pyramidal-CNN along with region embeddings. Even on the Elec, DBPedia, and RCV1 datasets, our results present significant improvements over the previous best semi-supervised results. $L_{\textit{MIXED}}$ objective also scales well to the dataset sizes, as on the large datasets of IMDB and Arxiv, it outperforms the above mentioned previous approaches by a substantial margin. We note here that the approach of~\citet{howard2018universal} is not directly comparable with our results as they use a three-layer LSTM model. We discuss the effect of model size in \S\ref{sec:analysis}.

\begin{table}[t]
\small
\centering
\begin{tabular}{@{} c c c c c c | c @{}} 
\toprule
$L$ & $U$ & $\lambda_{\textit{ML}}$ & $\lambda_{\textit{AT}}$ & $\lambda_{\textit{EM}}$ & $\lambda_{\textit{VAT}}$ & ACL-IMDB \\
\midrule
1 & 0 & 1 & 0 & 0 & 0 & 6.43 \\
1 & 0 & 1 & 1 & 0 & 0 & 5.96 \\
1 & 0 & 1 & 0 & 1 & 0 & 6.46 \\
1 & 0 & 1 & 0 & 0 & 1 & 5.98 \\
1 & 0 & 1 & 1 & 1 & 1 & 5.68 \\
\midrule
1 & 1 & 1 & 0 & 1 & 0 & 5.78 \\
1 & 1 & 1 & 0 & 0 & 1 & 5.52 \\
1 & 1 & 1 & 0 & 1 & 1 & 4.47 \\
1 & 1 & 1 & 1 & 1 & 1 & 4.32 \\
\bottomrule
\end{tabular}
\caption{Error rates (\%) for ablation study on the importance of hyperparameters when the BiLSTM model is trained using $L_{\textit{MIXED}}$ objective; $L$ = was labeled data used? $U$ = was unlabeled data used?}
\label{table:ablation_unsup}
\end{table}

Next, we perform ablation studies when the model is trained using $L_{\textit{MIXED}}$ on the ACL-IMDB dataset and analyze the contributions of the different component terms present in the objective (see Table~\ref{table:ablation_unsup}). First, we observe that when the model is trained using $L_{\textit{MIXED}}$ objective on both the labeled and unlabeled data, the accuracy on ACL-IMDB drastically improves by 33\% compared with using only the $L_{\textit{ML}}$ objective. Second, we also observe that when trained only on labeled data the inclusion of $L_{\textit{AT}}$ and $L_{\textit{VAT}}$ can also significantly improve the performance. However, $L_{\textit{EM}}$ alone doesn't lead to any significant gains. Furthermore, when $L_{\textit{MIXED}}$ is trained with only labeled data, we see 12\% relative increase in accuracy. Finally, when we add unlabeled data to both $L_{\textit{VAT}}$ and $L_{\textit{EM}}$, we see consistent improvements, thus suggesting that these objective functions complement each other and together improve the overall performance.

\section{Analysis}                        \label{sec:analysis}

\subsection{Effect on Word Embeddings}
\begin{table}[!htb]
\small
\centering
\begin{tabular}{@{}l | l | l@{}} 
\toprule
\multicolumn{3}{c}{\emph{query word}: good} \\
\midrule
\textbf{word2vec} & \textbf{ML Objective} & \textbf{Mixed Objective} \\ 
\midrule
great (0.64) & great (0.66) & funny (0.73) \\
really (0.61) & nice (0.62) & well-acted (0.72) \\
decent (0.60) & decent (0.58) & interesting (0.66) \\
nice (0.59) & entertaining (0.56) & fine (0.65) \\
ok (0.57) & overall (0.55) & nice (0.65) \\
but (0.56) & really (0.55) & thought-provoking (0.63) \\
pretty (0.56) & liked (0.54) & decent (0.62) \\
overall (0.55) & lot (0.52) & worth (0.60) \\
bad (0.54) & enjoyable (0.52) & recommend (0.60) \\
movie (0.53) & fun (0.51) & recommendable (0.60) \\
\bottomrule
\end{tabular}
\caption{Top-10 nearest neighbors according to cosine similarity (shown in parentheses) for the word ``\emph{good}'' computed in the embedding space. ``\emph{word2vec}'' refers to the static embeddings i.e.\ not finetuned during training.}
\label{table:t-sne}
\end{table}

To understand the effect on word embeddings due to training using $L_{\textit{ML}}$ and $L_{\textit{MIXED}}$ objectives, we show the top-10 closest words for the query word ``\emph{good}'' based on their cosine similarity in Table~\ref{table:t-sne}, where the word embeddings were extracted from the models trained on ACL-IMDB dataset. We see that for static embeddings, the closest words have a mix of both positive (`\emph{great}', `\emph{decent}', `\emph{nice}') and negative sentiments (`\emph{bad}', `\emph{but}'). This can be understood as they are syntactically similar adjectives. When these embeddings are finetuned using the $L_{\textit{ML}}$ objective, the network learns more meaningful representations and accommodates more positive sentiment words close to the query word `\emph{good}'. Moreover, when trained using the $L_{\textit{MIXED}}$ objective, we see that those words that have a very high correlation with the class label (\emph{positive sentiment} class in this case) are clustered together in the embedding space. Our hypothesis is that this factor also contributes to an increase in the overall classification accuracy.

\subsection{Model Regularization Effect}
\begin{figure}[!htp]
\begin{minipage}{0.5\linewidth}
\centering
\includegraphics[scale=0.27]{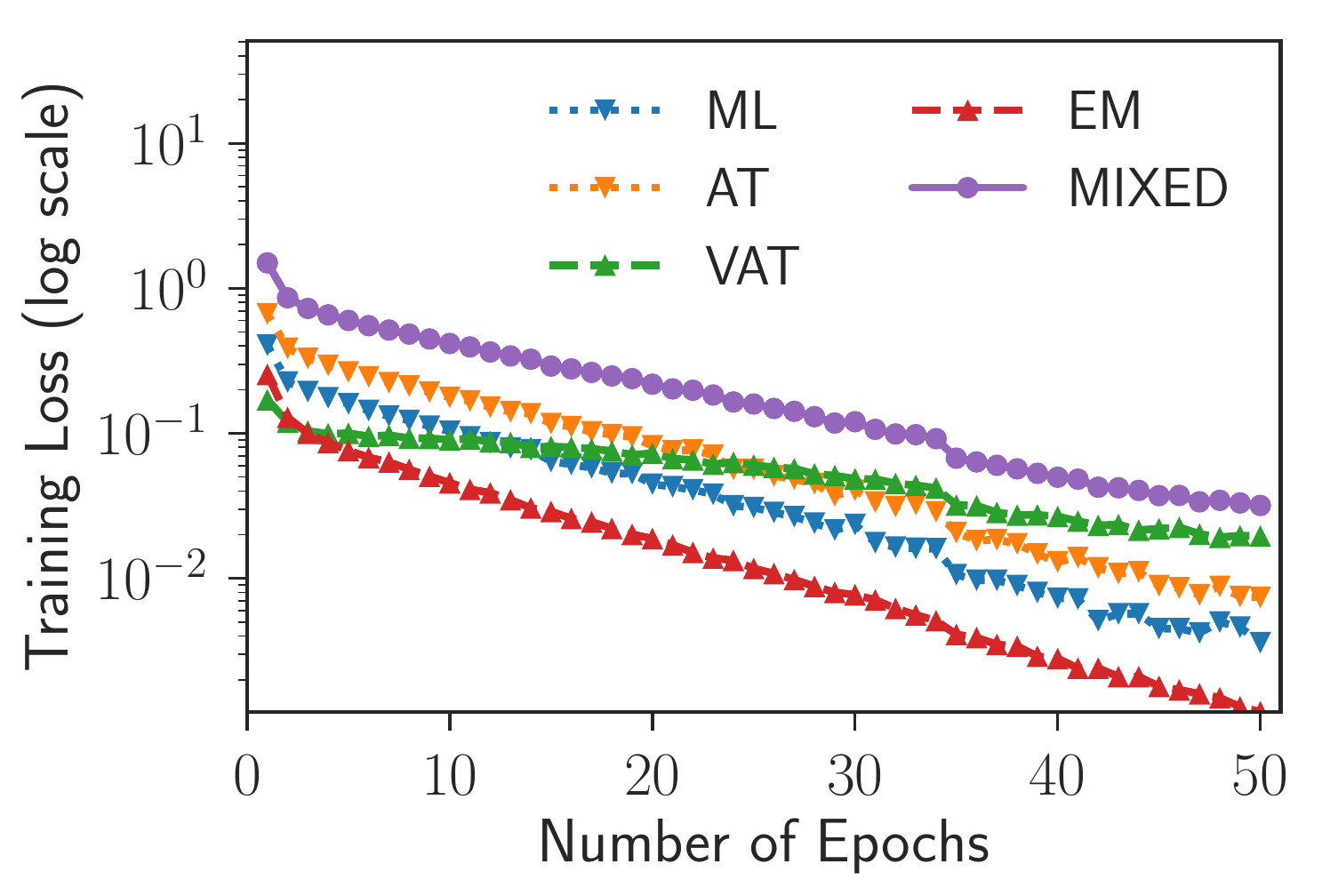}
\end{minipage}%
\begin{minipage}{0.5\linewidth}
\centering
\includegraphics[scale=0.27]{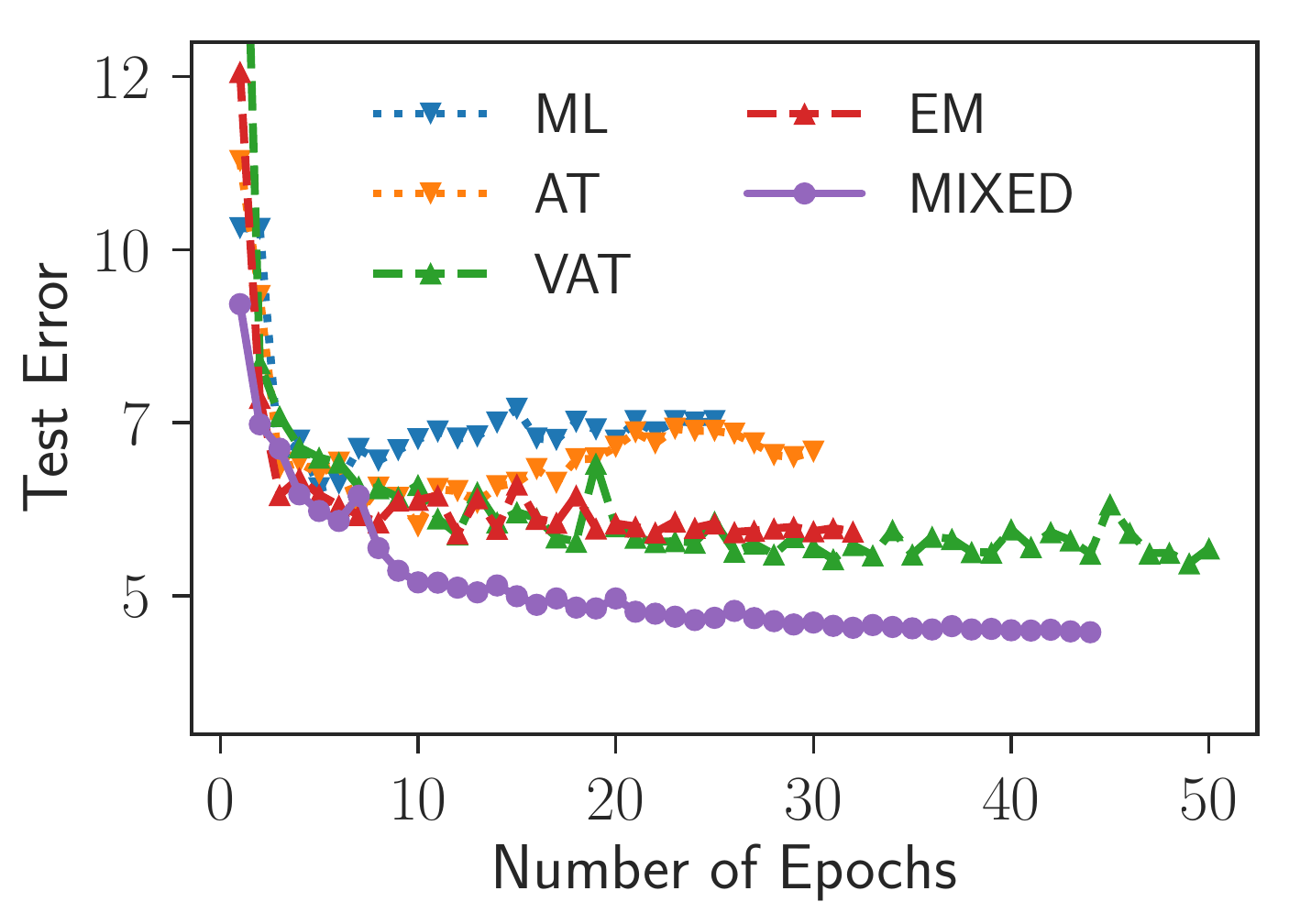}
\end{minipage}
\caption{(a) Training loss vs.\ epochs on ACL-IMDB; (b) Test error vs.\ epochs on ACL-IMDB.}
\label{fig:reg}
\end{figure}

Figure~\ref{fig:reg}a and Figure~\ref{fig:reg}b show the moving average training loss and test error respectively versus the number of epochs on the ACL-IMDB dataset with the $L_{\textit{ML}}$, $L_{\textit{AT}}$, $L_{\textit{VAT}}$, $L_{\textit{EM}}$, and $L_{\textit{MIXED}}$ objectives. We can see
 that $L_{\textit{ML}}$ begins to overfit after 5 epochs, $L_{\textit{AT}}$ overfits after 10 epochs while $L_{\textit{MIXED}}$, $L_{\textit{EM}}$, and $L_{\textit{VAT}}$ don't overfit much and thus achieve better generalization than the $L_{\textit{ML}}$ and $L_{\textit{AT}}$ objectives (see in Figure~\ref{fig:reg}b). Moreover, as $L_{\textit{MIXED}}$ and $L_{\textit{VAT}}$ objectives can use unlabeled data, their training loss decays gradually.
Thus, $L_{\textit{MIXED}}$ objective while being very effective in performance is also a very robust model regularizer. On the other hand, from Figure~\ref{fig:reg}b, we can see that $L_{\textit{VAT}}$, $L_{\textit{EM}}$, and $L_{\textit{MIXED}}$ take a long time to converge compared with $L_{\textit{ML}}$ and $L_{\textit{AT}}$ and are thus quite slow to train. In our experiments, one epoch of $L_{\textit{MIXED}}$ takes around 20m on GeForce GTX 1080 GPU and it requires roughly 45 epochs to converge. This is considerably slower than $L_{\textit{ML}}$ objective where each epoch takes around 3m and the overall convergence time is thus 15m for 5 epochs.

\subsection{Varying Data Size}
\begin{figure}[!htp]
\begin{minipage}{0.5\linewidth}
\centering
\includegraphics[scale=0.29]{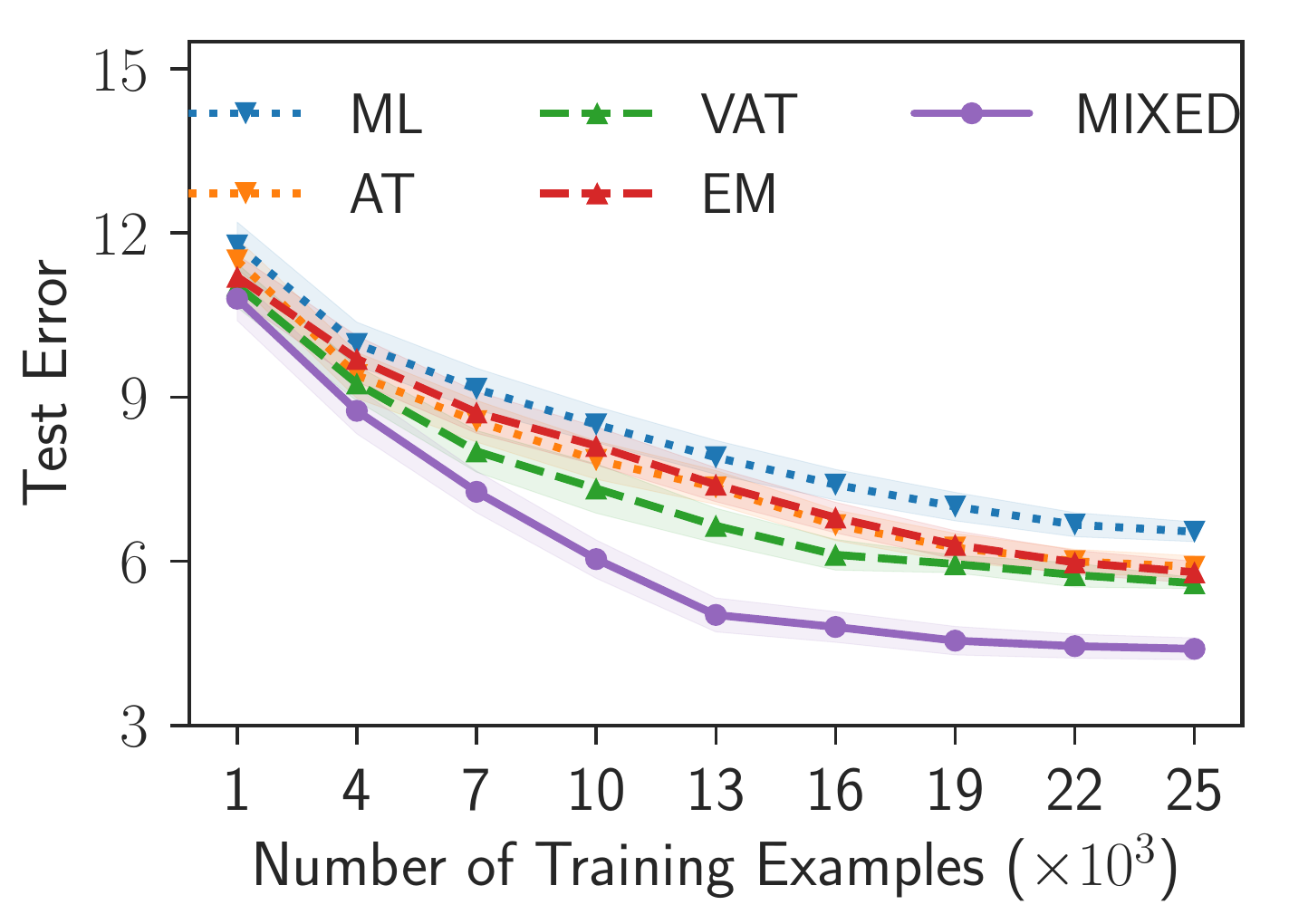}
\end{minipage}%
\begin{minipage}{0.5\linewidth}
\centering
\includegraphics[scale=0.29]{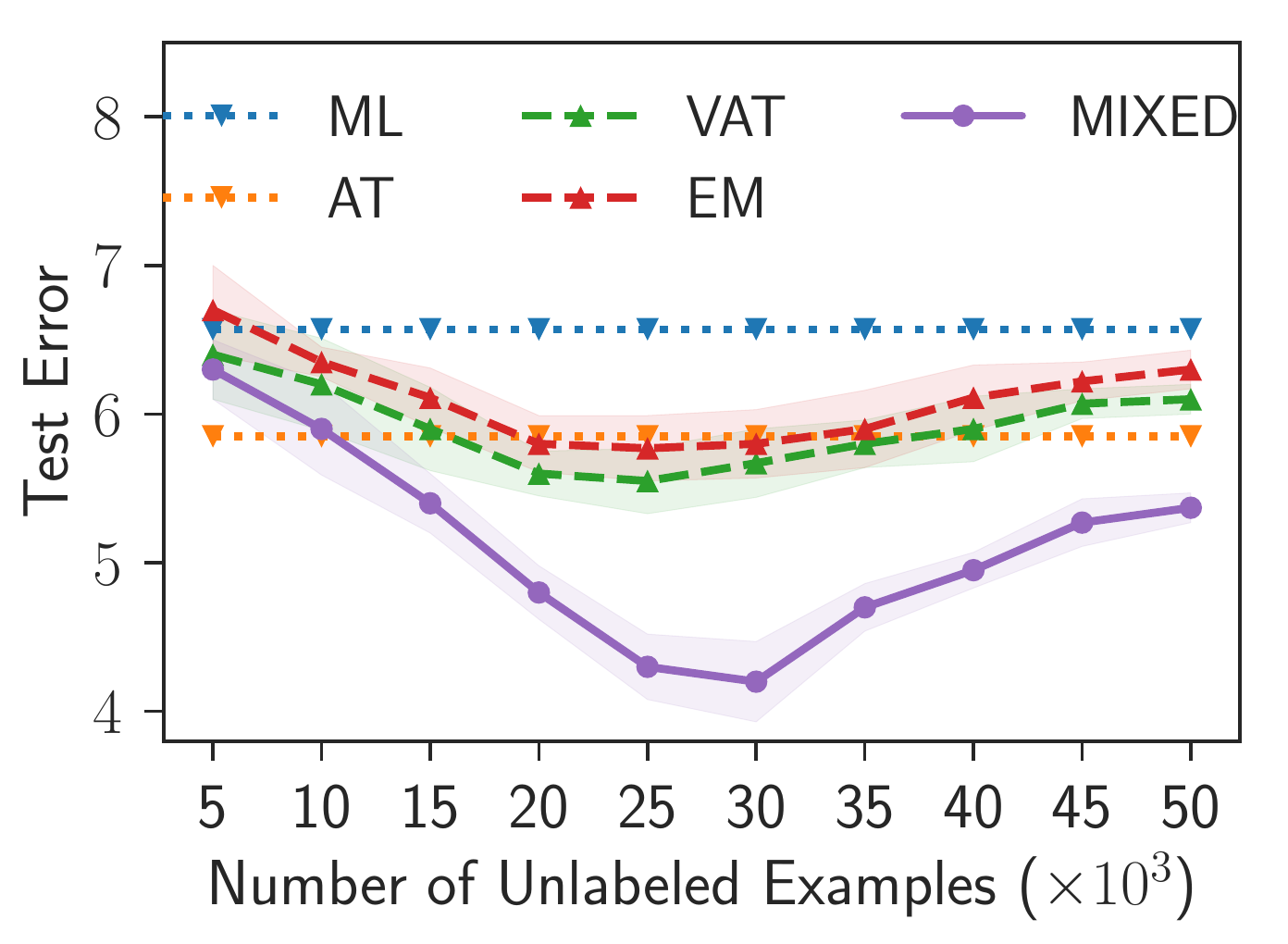}
\end{minipage}
\caption{Test Error on ACL-IMDB vs. (a) number of training examples (b) increasing number of unlabeled examples.}
\label{fig:varying_data}
\end{figure}

In this setup, we first analyze the test error on ACL-IMDB dataset by feeding the model trained with different objective functions (\S\ref{sec:methods}) with an increasing number of training examples (learning curve; see Figure~\ref{fig:varying_data}a). We observe that all the objective functions converge to lower error rates when training data is increased. We also see that mixed objective model is always optimal (achieves lower test error rate) for any setting of the number of training examples.

Next, we analyze the test error on ACL-IMDB dataset by varying the amount of unlabeled data. For this experiment, we use additional 50,000 reviews provided with the ACL-IMDB dataset and its 25,000 reviews from test set as unlabeled data. We evaluate the performance of each objective function by linearly increasing the amounts of unlabeled data (see Figure~\ref{fig:varying_data}b). Initially, increasing the amount of unlabeled data tends to improve the performance of $L_{\textit{VAT}}$, $L_{\textit{EM}}$, and $L_{\textit{MIXED}}$. However, we observe that their performance saturates once 25,000 unlabeled examples are available. Furthermore, as the amount of unlabeled data increases, the performance tends to degrade sharply. As ACL-IMDB training set also consists of 25,000 examples, from this observation, it can be assumed that to obtain the best performance using $L_{\textit{MIXED}}$, the size of unlabeled and labeled dataset should be roughly the same. We also note that as $L_{\textit{ML}}$ and $L_{\textit{AT}}$ are supervised approaches, their performance remains unaffected.

\begin{table*}[t]
\begin{minipage}{\linewidth}
\small
\centering
\begin{tabularx}{\linewidth}{@{} l X @{}}
\toprule
\textbf{Sentiment} & \textbf{Text} \\
\midrule
\multicolumn{2}{c}{\textbf{Mixed Objective Training} ($L_{\textit{MIXED}}$)} \\
\midrule
Negative & This movie is so over-the-top as to be a borderline comedy. Laws of physics are broken. Things explode for no good reason. Great movie to sit down with a six-pack and enjoy. Do not - I repeat DO NOT see this movie sober. You will die horrible death!\\
\midrule
\multicolumn{2}{c}{\textbf{Entropy Minimization Training} ($L_{\textit{EM}}$)} \\
\midrule
Positive & Coming from Oz I probably shouldn't say it but I find a lot of the local movies lacking that cohesive flow with a weak storyline. This comedy lacks in nothing. Great story, no overacting, no melodrama, just brilliant comedy as we know Oz can do it. Do yourself a favour and laugh till you drop.\\
\midrule
\multicolumn{2}{c}{\textbf{Virtual Adversarial Training} ($L_{\textit{VAT}}$)} \\
\midrule
Negative & The plot seemed to be interesting, but this film is a great dissapointment. Bad actors, a camera moving like in the hands of an amateur. If there was C-movies, this would be a perfect example. A plus for a nice DVD cover though and a great looking female actor.\\
% \midrule
% \multicolumn{2}{c}{\textbf{Adversarial Training} ($L_{\textit{AT}}$)} \\
% \midrule
% Positive & Low budget, mostly no name actors ... this is what a campy horror flick is supposed to be all about. These are the types of movies that kept me on the edge of my seat as a kid staying up too late to watch cable. If you liked the 80's horror scene this is the movie for you. \\
% \midrule
% \multicolumn{2}{c}{\textbf{Maximum Likelihood Training} ($L_{\textit{ML}}$)} \\
% \midrule
% Negative & OK, so it's a silly movie, but I think they knew that when they made it. And there are some neat little twists on the otherwise tired, overdone ``Godzilla''-type genre. Borrowed a tape just because I knew someone in it, but I did loan it out to a couple pals, who also kinda liked it. \\
\bottomrule
\end{tabularx}
\caption{Examples from ACL-IMDB dataset for sentiment classification task that are correctly classified by the method indicated directly above it and incorrectly classified by all the other methods.}
\label{table:examples}
\end{minipage}
\end{table*}

\subsection{Varying Model Size}
\begin{figure}[!htp]
\begin{minipage}{0.5\linewidth}
\centering
\includegraphics[scale=0.29]{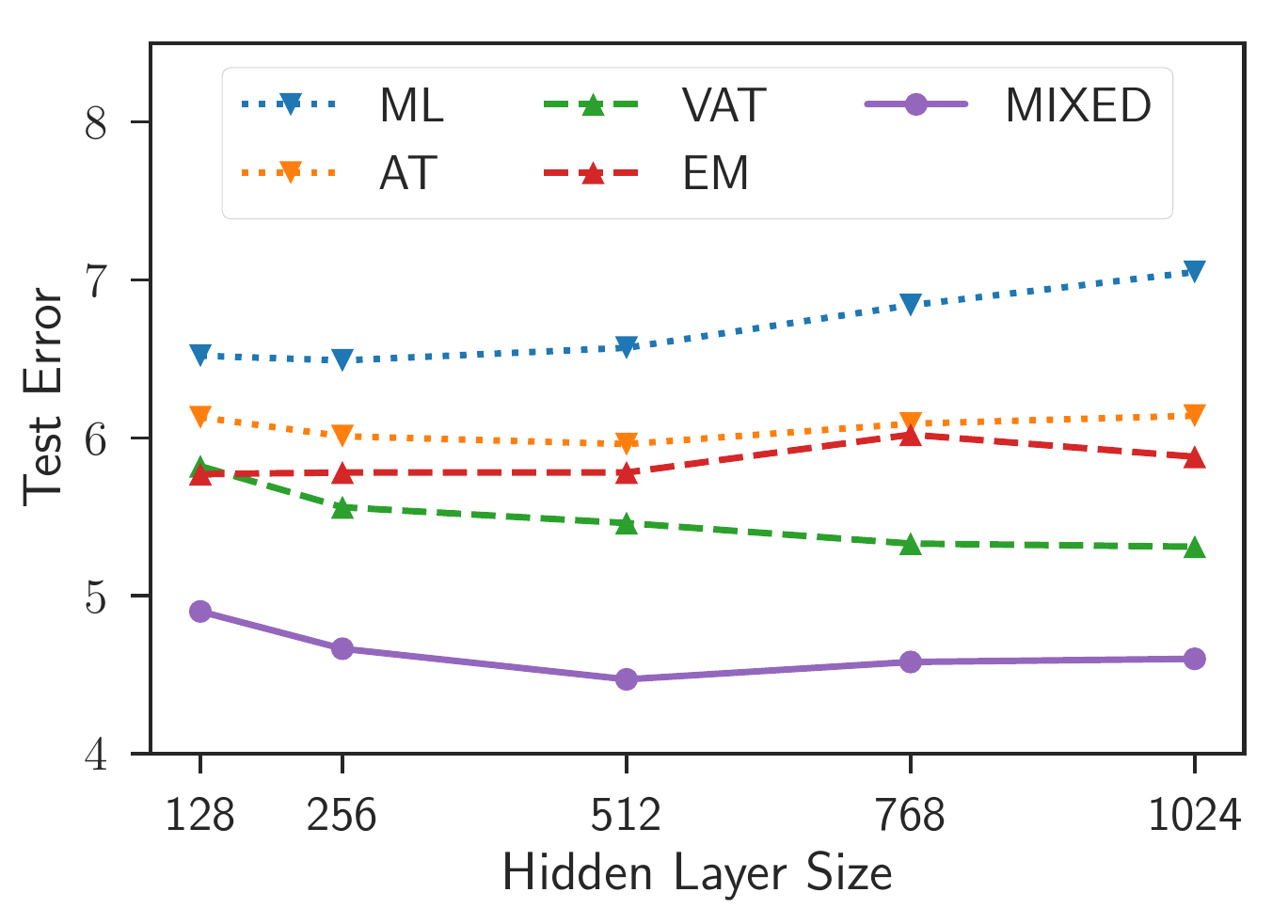}
\label{fig:hidden_size}
\end{minipage}%
\begin{minipage}{0.5\linewidth}
\centering
\includegraphics[scale=0.29]{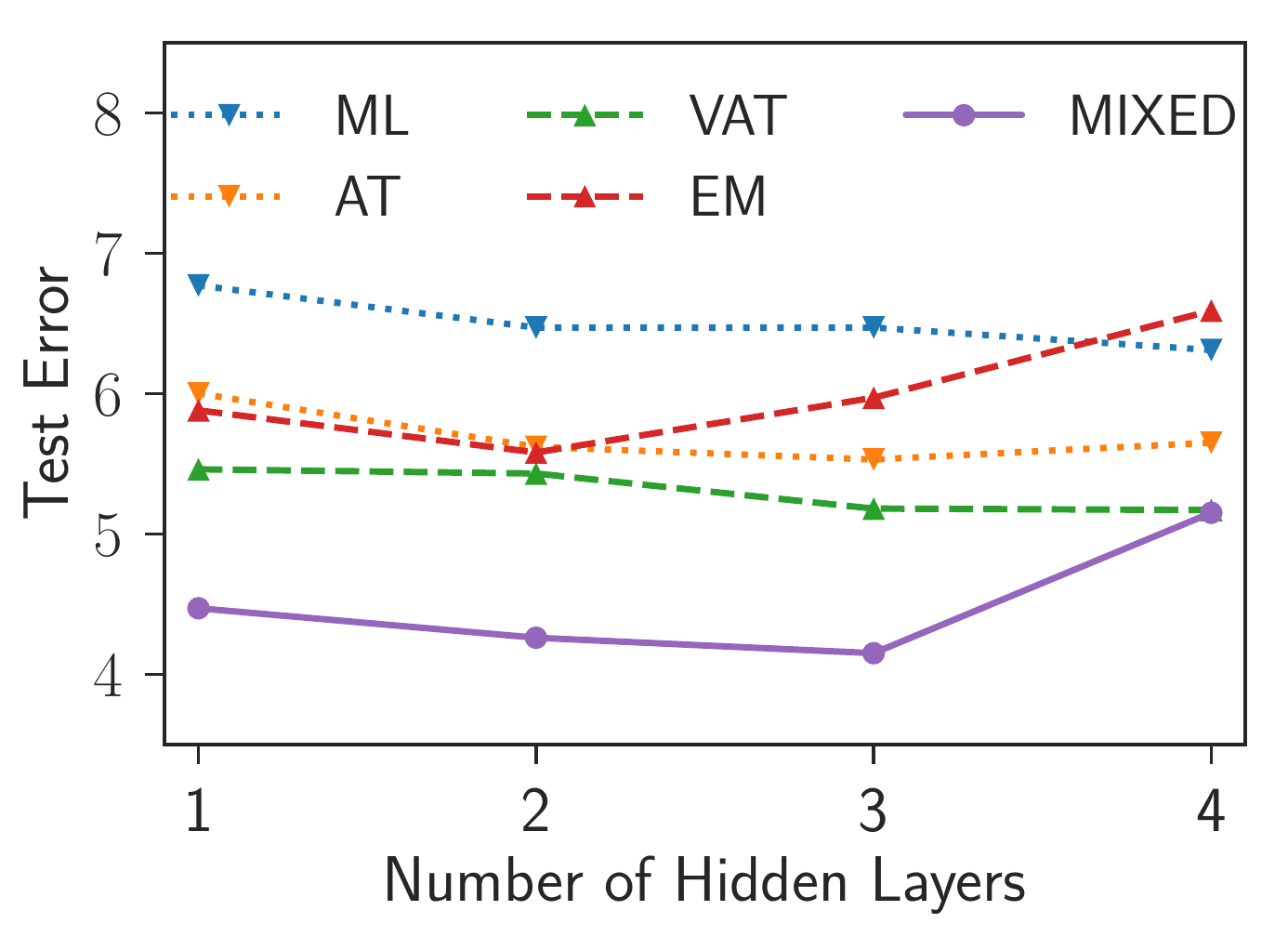}
\label{fig:layers}
\end{minipage}
\caption{Test error on ACL-IMDB vs. (a) hidden layer size of LSTMs in the BiLSTM with one-hidden layer; (b) number of BiLSTM layers where each LSTM has size of 512.}
\label{fig:model_size}
\end{figure}

We note that prior work in the supervised text classification task has used smaller-sized LSTMs \emph{i.e.}\ models with hidden state sizes at most 512 units because larger models didn't give accuracy gains (DL15, JZ16). This is also consistent with our observation, as we find that that supervised approaches ($L_{\textit{ML}}$) do not benefit much from increasing the model size. However, when using additional loss functions such as the mixed objective, accuracy scales much better with model size (see Figure~\ref{fig:model_size}a). Further, we also observe accuracy gains for all methods upon increasing the number of layers in the model (see Figure~\ref{fig:model_size}b). Specifically, the error rate of $L_{\textit{MIXED}}$ objective improves to 4.15\% when using a three-layer deep model. This suggests that larger-sized semi-supervised methods can lead to the development of more accurate models for text classification task. However, a four-layer model hurts the $L_{\textit{MIXED}}$ objective's performance due to the training instability of $L_{\textit{EM}}$ method.

\subsection{Effect on Prediction Probabilities}
\begin{figure}[t]
\begin{minipage}{\linewidth}
\centering
\includegraphics[width=\linewidth]{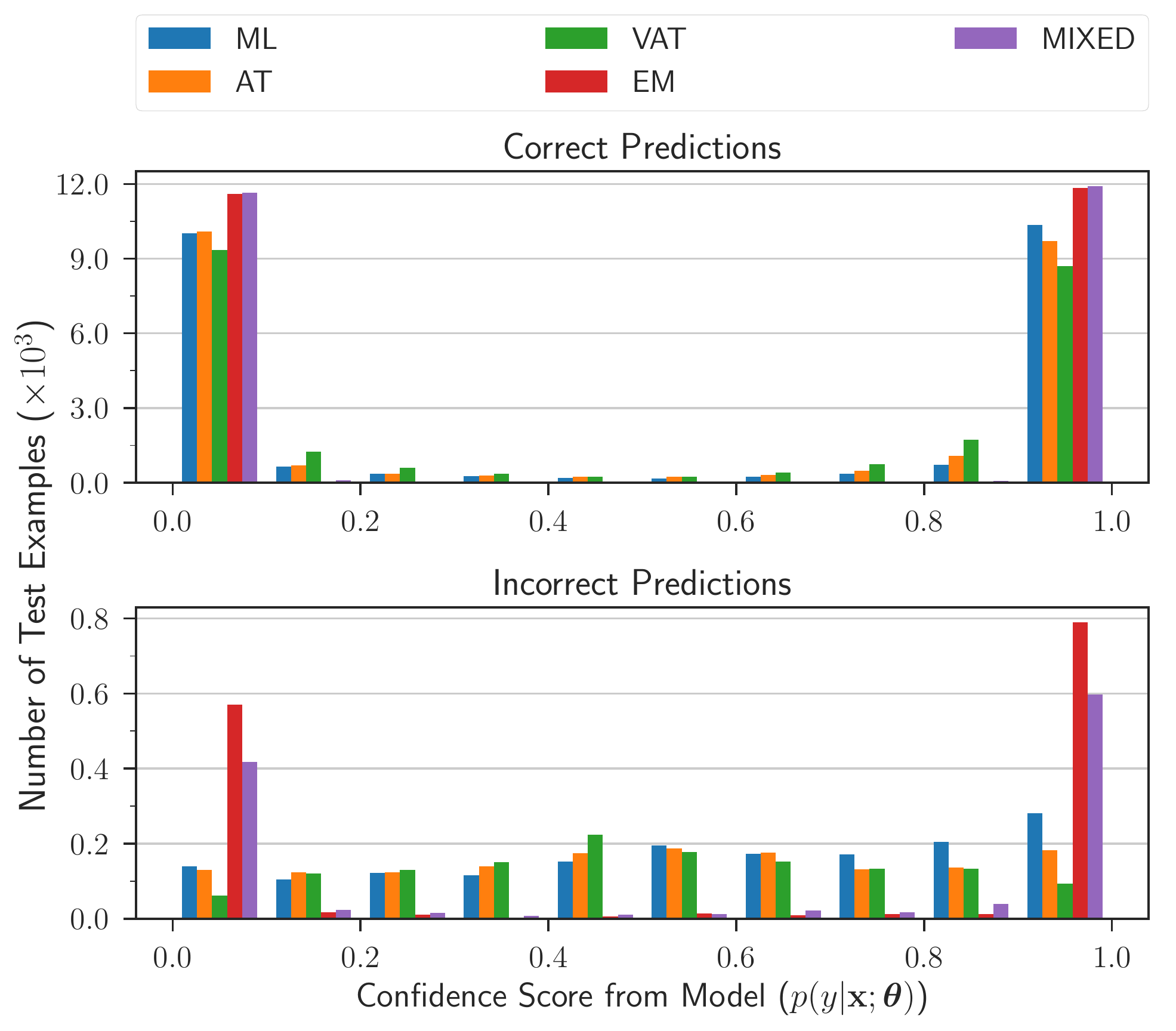}
\end{minipage}
\caption{Histogram of prediction probabilities on ACL-IMDB test set for the case of (a) correct predictions; (b) incorrect predictions.}
\label{fig:model_histogram}
\end{figure}

To study the behavior of different methods, we plot a histogram of the prediction probabilities for both the correct (Figure~\ref{fig:model_histogram}a) and incorrect (Figure~\ref{fig:model_histogram}b) predictions.
We observe that for correct predictions all the methods especially $L_{\textit{EM}}$ and $L_{\textit{MIXED}}$ have very sharp and confident distribution of class probabilities. However, for incorrect predictions only $L_{\textit{EM}}$ has sharp peaks while $L_{\textit{VAT}}$, $L_{\textit{AT}}$, and $L_{\textit{ML}}$ encourage the model to learn a smoother distribution.

\subsection{Ensemble Approach vs Mixed Objective}
To understand if the above objectives have complementary strengths, we combine their predicted probabilities with a linear interpolation strategy. Given the output probability for a class $k$ as $p(y=k|\mathbf{x})$, the interpolated probability $p_{\textit{I}}(y=k|\mathbf{x})$ is calculated as:
\begin{align*}
p_{\textit{I}} = \alpha_{\textit{ML}} p_{\textit{ML}} + \alpha_{\textit{AT}} p_{\textit{AT}} + \alpha_{\textit{VAT}} p_{\textit{VAT}} + \alpha_{\textit{EM}} p_{\textit{EM}} \mid \sum\alpha_{\textit{i}} = 1,
\end{align*}
where $\alpha_{\textit{i}} \in [0, 1]$ and is chosen based on grid search. This simple interpolation technique results in an improved error rate of 5.2\%. However, the error rate of our proposed mixed objective function is substantially lower (4.3\%) thus highlighting the importance of performing joint training of the model based on different objective functions.

\subsection{Example Predictions}
In Table~\ref{table:examples}, we show some example movie reviews from the test set of ACL-IMDB dataset that are correctly classified by the highlighted method and incorrectly by all the remaining methods. We observe that methods such as $L_{\textit{MIXED}}$, $L_{\textit{EM}}$, and $L_{\textit{VAT}}$ that are based on unsupervised training are able to correctly classify difficult instances in which the overall sentiment is determined by the entire sentence structure. This illustrates the ability of these methods to learn complex long-range dependencies.

\section{Relation Extraction}    \label{sec:RE}
To evaluate if the mixed objective function can generalize to other tasks we also perform experiments on relation extraction (RE) task. In this, the objective is to identify if a predefined semantic ``\emph{relation category}'' exists (or doesn't exist) between a pair of \emph{subject} and \emph{object} entities present in text. This task also presents specific challenges: the linguistics \emph{coverage problem} due to the lack of all possible training examples of a relation class and longer text span between the entities in a sentence.

For the RE task, we use a position-aware attention model~\citep{yuhao2017slot} that consists of a word embedding, position embedding, LSTM, attention layer, linear layer, and softmax layer. We augment the word embeddings by concatenating them with POS tag and NER category embeddings which is then fed to the LSTM to get hidden states for each word. The position embeddings for a word is derived based on the relative distance of the current word from the subject and object entities. Next, the attention layer computes the final sentence representation by focusing on both the hidden states and position embeddings. Finally, sentence representation is fed to the linear layer followed by a softmax layer for relation classification.

We perform experiments using our proposed mixed objective function on two RE datasets: \emph{TACRED} and \emph{SemEval-2010 Task 8} whose statistics are shown in Table~\ref{table:RE_dataset_stat}. The POS and NER tags are computed using the Stanford CoreNLP toolkit.\footnote{\corenlpurl} Following standard convention, we report the micro-averaged F\textsubscript{1} score on TACRED and official macro-averaged F\textsubscript{1} score on SemEval datasets. We performed only a small number of experiments to search for the hyper-parameter values of dropout, embedding size, hidden layer size, and learning rate on the development set, all other parameters remained the same as the positional attention model of~\citet{yuhao2017slot}.\footnote{\parelexturl} Our results in Table~\ref{table:RE_accuracy} show that when trained with mixed objective function, our model performs quite well, producing better results than all previously reported models despite the lack of complex task-specific hyper-parameter tuning.
\begin{table}[t]
\begin{minipage}{\linewidth}
\small
\centering
\begin{tabular}{@{} l | r r r r r r @{}}
 \toprule
 \textbf{Dataset} & \textbf{Train} & \textbf{Dev} & \textbf{Test} & \textbf{\% Neg} & \textit{\textbf{K}} & $\boldsymbol{\ell}$\\
 \midrule
 TACRED & 68,124 & 22,631 & 15,509 & 79.5\% & 42 & 36 \\
 SemEval & 8,000 & \multicolumn{1}{c}{--} & 2,717 & 17.4\% & 19 & 19 \\
 \bottomrule
 \end{tabular}
\caption{Summary statistics for RE datasets; \% Neg = percentage of examples with class label of ``\emph{no relation}'' between entities; \textit{K} = number of classes including the \emph{no relation} class; \textit{$\ell$} = average length of a sentence in the dataset.}
\label{table:RE_dataset_stat}
\medskip
\end{minipage}
\begin{minipage}{\linewidth}
\small
\centering
\begin{tabular}{@{} l | c c c | c c c @{}}
 \toprule
  & \multicolumn{3}{c|}{\textbf{TACRED}} & \multicolumn{3}{|c}{\textbf{SemEval-2010}} \\
 \cmidrule{2-7}
 \textbf{Model} & \textbf{P} & \textbf{R} & \textbf{F\textsubscript{1}} & \textbf{P} & \textbf{R} & \textbf{F\textsubscript{1}} \\
 \midrule
 CNN-PE\textsuperscript{a} & \textbf{70.3} & 54.2 & 61.2 & 82.1 & 83.1 & 82.5 \\
 SDP-LSTM\textsuperscript{b} & 66.3 & 52.7 & 58.7 & -- & -- & 83.7 \\
 PA-LSTM\textsuperscript{c} & 65.7 & 64.5 & 65.1 & -- & -- & 82.7 \\
 \midrule
Our Method & 66.4 & \textbf{67.3} & \textbf{66.8} & \textbf{83.5} & \textbf{84.8} & \textbf{84.1} \\ 
 \bottomrule
 \end{tabular}
\caption{Model performance on TACRED and SemEval datasets; P = Precision; R = Recall; \textsuperscript{a}\citet{santos2015relation}; \textsuperscript{b}\citet{yan2015relation};
\textsuperscript{c}\citet{yuhao2017slot}.
For fair comparison, we only report results from models that don't use ensembling or ranking-based approaches.}
\label{table:RE_accuracy}
\end{minipage}
\end{table}

\section{Related Work}               \label{sec:related_work}

\subsubsection*{Neural Network Methods.}
Neural network models for NLP have yielded impressive results on several benchmark tasks~\citep{collobert2011natural,peters2018deep}. To learn document features for text classification task, several methods have been proposed---~\citet{kim2014convolutional} uses 1D CNNs,~\citet{lai2015recurrent} uses a simple bidirectional recurrent CNN with max-pooling,~\citet{zhou2016text} applies 2D max-pooling on top of BiLSTMs,~\citet{Zhou2015clstm} investigates a joint CNN-LSTM model, and JZ15a, JZ16, JZ17 apply CNNs, LSTMs, and pyramidal CNNs respectively to one-hot encoding of word sequences. An alternative approach is to first learn sentence representations followed by combining them to learn document features. To do this,~\citet{tang2015document} first apply a CNN or LSTM followed by a gated RNN while~\citet{yang2016hierarchical} learn the sentence and document features in a hierarchical manner using a self-attention mechanism.

\subsubsection*{Semi-Supervised Learning.} SSL approaches can be broadly categorized into three types: multi-view, data augmentation, and transfer learning. 
First, under multi-view learning, the objective is to use multiple views of both the labeled and unlabeled data to train the model. These multiple views can be obtained either from raw text~\citep{blum1998combining} or from the features (JZ15b). Second, under data augmentation, as the name implies, involves pseudo-augmenting either the features or the labels. For text classification,~\citet{nigam2000text} performed semi-supervised training using na\"ive Bayes and expectation-maximization algorithms and demonstrated substantial improvements in performance. MDG16 compute embedding perturbations using adversarial and virtual adversarial approaches to improve model training. Third, under transfer learning, the approach of initializing the task-specific model weights by pretrained weights from an auxiliary task is a widely used strategy that has shown to improve the performance in tasks such as text classification~\citep{dai2015semi,howard2018universal}, question-answering~\citep{devlin2018bert}, and machine translation~\citep{ramachandran2017pretraining,qi2018embeddings}.

\section{Conclusion}   \label{sec:conclusion}
We show that a simple BiLSTM model using maximum likelihood training can result in a competitive performance on text classification tasks without the need for an additional pretraining step. Also, in addition to maximum likelihood, using a combination of entropy minimization, adversarial, and virtual adversarial training, we report state-of-the-art results on several text classification datasets. This mixed objective function also generalizes well to other tasks such as relation extraction where it outperforms current best models.

\small
\bibliography{aaai}
\bibliographystyle{aaai}
\end{document}